%% file: a_TAI_arxiv_submission.tex
\documentclass[journal]{IEEEtai}

\usepackage[colorlinks,urlcolor=blue,linkcolor=blue,citecolor=blue]{hyperref}

\usepackage{color,array}

\usepackage{amsmath,amssymb,amsfonts}
\usepackage{algorithmic}
\usepackage{graphicx}
\usepackage{textcomp}
\usepackage{algorithm}
\usepackage{multirow}
\usepackage{graphicx}

\usepackage{bm}
\newcommand{\mathbold}[1]{\bm{#1}}

\begin{document}

\title{CALR: Corrective Adaptive Low-Rank Decomposition for Efficient Large Language Model Layer Compression}

\author{Muchammad Daniyal Kautsar, Afra Majida Hariono, Widyawan, Syukron Abu Ishaq Alfarozi*, \IEEEmembership{Member, IEEE}, Kuntpong Woraratpanya, \IEEEmembership{Member, IEEE}
% \thanks{This paragraph of the first footnote will contain the date on which you submitted your paper for review. It will also contain support information, including sponsor and financial support acknowledgment. For example, ``This work was supported in part by the U.S. Department of Commerce under Grant BS123456.'' }
\thanks{Muchammad Daniyal Kautsar, Widyawan, Afra Majida Hariono, and Syukron Abu Ishaq Alfarozi are with the Department of Electrical and Information Engineering, Universitas Gadjah Mada, Yogyakarta 55281, Indonesia (e-mail: muchammad.daniyal.kautsar@mail.ugm.ac.id, widyawan@ugm.ac.id, afra.majida0202@mail.ugm.ac.id, and syukron.abu@ugm.ac.id).}
\thanks{Kuntpong Woraratpanya is with the School of Information Technology, King Mongkut’s Institute of Technology Ladkrabang, Bangkok 10520, Thailand (e-mail: kuntpong@it.kmitl.ac.th).}

\thanks{*Corresponding Author: Syukron Abu Ishaq Alfarozi (syukron.abu@ugm.ac.id).}

\thanks{This work has been submitted to the IEEE for possible publication. Copyright may be transferred without notice, after which this version may no longer be accessible.}
}

\markboth{Journal of IEEE Transactions on Artificial Intelligence, Vol. 00, No. 0, Month 2020}
{Muchammad Daniyal Kautsar \MakeLowercase{\textit{et al.}}: Bare Demo of IEEEtai.cls for IEEE Journals of IEEE Transactions on Artificial Intelligence}

\maketitle

\begin{abstract}
\input{02-abstract}
\end{abstract}

\begin{IEEEImpStatement}
The growing adoption of Large Language Models (LLMs) is limited by their enormous size and heavy computational requirements. These constraints create high economic barriers for smaller organizations and raise environmental concerns due to substantial energy use. Our research proposes a compression method that not only reduces the size of these models but also learns to recover essential functional information that is usually lost during compression. Unlike traditional techniques that only approximate mathematical structures, our approach focuses on preserving the model’s ability to reason and perform tasks effectively. As a result, powerful AI systems can run on everyday devices, such as laptops and smartphones. This significantly reduces deployment costs and energy consumption. By making advanced AI more accessible and efficient, our work promotes wider innovation and supports the development of privacy-friendly, on-device AI applications that benefit society.
\end{IEEEImpStatement}


% \section{Introduction}

\input{01-content-short}

\bibliographystyle{IEEEtran}
\bibliography{references}

\end{document}

%% file: 02-abstract.tex
Large Language Models (LLMs) present significant deployment challenges due to their immense size and computational requirements. Model compression techniques are essential for making these models practical for resource-constrained environments. A prominent compression strategy is low-rank factorization via Singular Value Decomposition (SVD) to reduce model parameters by approximating weight matrices. However, standard SVD focuses on minimizing matrix reconstruction error, often leading to a substantial loss of the model's functional performance. This performance degradation occurs because existing methods do not adequately correct for the functional information lost during compression. To address this gap, we introduce Corrective Adaptive Low-Rank Decomposition (CALR), a two-component compression approach. CALR combines a primary path of SVD-compressed layers with a parallel, learnable, low-rank corrective module that is explicitly trained to recover the functional residual error. Our experimental evaluation on SmolLM2-135M, Qwen3-0.6B, and Llama-3.2-1B, demonstrates that CALR can reduce parameter counts by 26.93\% to 51.77\% while retaining 59.45\% to 90.42\% of the original model's performance, consistently outperforming LaCo, ShortGPT, and LoSparse. CALR's success shows that treating functional information loss as a learnable signal is a highly effective compression paradigm. This approach enables the creation of significantly smaller, more efficient LLMs, advancing their accessibility and practical deployment in real-world applications.

%% file: 01-content-short.tex
\section{Introduction}
\label{sec:introduction}

\PARstart{L}{arge} Language Models (LLMs) have revolutionized natural language processing, excelling in tasks from translation to the generation of creative content \cite{brown2020language, openai2023gpt4, hagos2024}. However, their immense size, encompassing billions to trillions of parameters, poses practical challenges, including demanding hardware requirements, high energy consumption, and limited deployment on resource-constrained devices \cite{zhu2024surveycompression, wang2024surveyslm}. These limitations motivate the development of efficient compression techniques that reduce model size while preserving performance.

\begin{figure*}[t!]
  \centering
  \includegraphics[width=\textwidth]{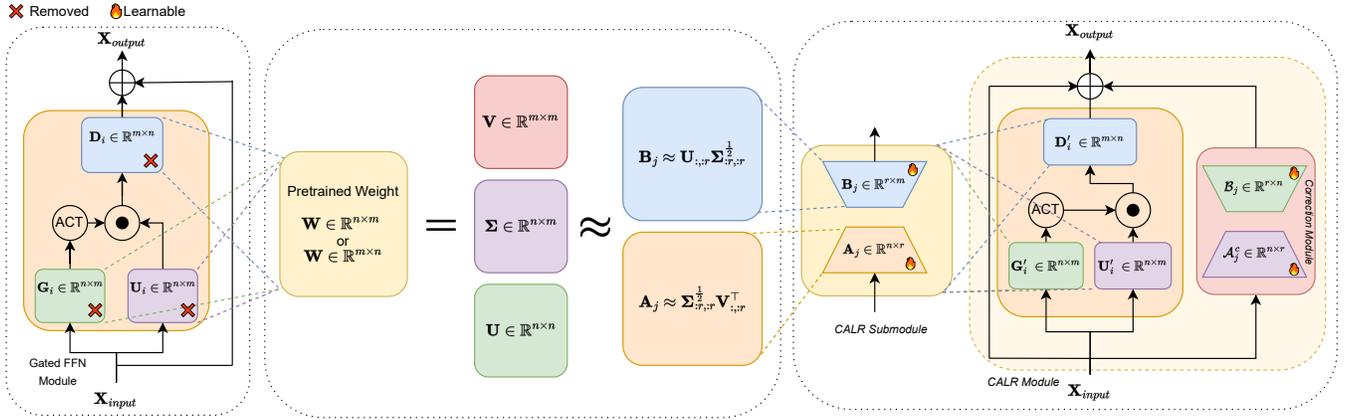}
  \caption{Conceptual overview of the CALR architectural transformation for a Gated FFN Module. The original pre-trained weights ($\mathbf{G}_i, \mathbf{U}_i, \mathbf{D}_i$) of the linear layers within the FFN module are replaced by their respective SVD-initialized low-rank factorizations ($\mathbf{A}_j, \mathbf{B}_j$, forming CALR submodules, denoted as $\mathbf{G}_i^{'}, \mathbf{U}_i^{'}, \mathbf{D}_i^{'}$). A separate, learnable, low-rank Correction Module ($\mathbold{\mathcal{A}}_j, \mathbold{\mathcal{B}}_j$) takes the same input as the FFN module ($\mathbf{X}_{\text{input}}$), and its output is added to the output of the SVD-compressed FFN path before the final residual connection.}
    \label{fig:calr_architecture}
\end{figure*}

Among various compression techniques, including pruning and quantization \cite{cheng2024surveypruning, fontana2024}, low-rank factorization is a prominent one. This technique allows for approximating large weight matrices through the product of smaller matrices, reducing the parameters while maintaining a differentiable structure compatible with modern hardware \cite{wang2025svdllm, kaushal2021lord}. Singular Value Decomposition (SVD) is a foundational tool for this task, as it provides the optimal low-rank approximation for a matrix in the Frobenius norm sense \cite{Eckart_Young_1936, MIRSKY1960, GOLUB1987317}. However, while attaining mathematical optimality in matrix reconstruction, this method is not guaranteed to maintain the functional effectiveness of a large language model (LLM).

Recent research has sought to enhance the application of SVD to LLMs. Methods like SVD-LLM \cite{wang2025svdllm} and ASVD \cite{anonymous2025asvd} introduce truncation-aware whitening and activation-aware transformations, respectively, while others like LoRD \cite{kaushal2021lord} and Lillama \cite{sy2025lillama} incorporate data-aware decomposition and feature distillation. A common thread in these works is optimizing a \textit{single} low-rank approximation for each weight matrix. While these methods improve the fidelity of this approximation, they are still bound by the expressive limits of a single low-rank component.

A recent work acknowledges this limitation by explicitly modeling the compression residual. An example is the ResSVD \cite{bai2025ressvd} method that performs a secondary SVD of the residual weight, while EoRA \cite{eora2024shih} introduces a low-rank with training-free pathway. Conceptually, the closest related framework is LoSparse \cite{li2023losparse}, which decomposes weights as $\mathbf{W} \approx \mathbf{U}\mathbf{V} + \mathbf{S}$, where $\mathbf{U}\mathbf{V}$ is the low-rank and $\mathbf{S}$ is the corrective \textit{sparse} matrix. LoSparse suggests that a sparse structure best describes the residual information. In contrast, this research is built on a different hypothesis: that is, for decoder-only LLMs, the functionally critical residual information is not sparse and isolated but distributed, and is better captured by a \textit{dense, low-rank} corrective structure, similar to how LoRA captures fine-tuning updates \cite{hu2021lora}.

To address this functional-mathematical disconnect, we introduce Corrective Adaptive Low-Rank Decomposition (CALR). CALR is distinguished by its twofold approach. First, it selectively targets specific, functionally ``quieter" layers for compression, with the acknowledgment that a uniform approach is suboptimal. Second, for these targeted layers, it employs a dual-path approach:

\begin{enumerate}
    \item Each target linear layer is replaced with its SVD-initialized low-rank factors, forming the primary compressed path.
    \item A parallel, additive, and learnable low-rank corrective module is introduced, trained explicitly to model the \textit{functional} error between the compressed and original module outputs.
\end{enumerate}

Fig. \ref{fig:calr_architecture} illustrates this framework. The corrective path processes the same input as the compressed primary path, and its output is added back, effectively compensating for SVD-induced functional loss. This approach positions CALR as a method for fine-grained, intra-module compression. It contrasts with structured pruning techniques that remove entire components, such as layers in ShortGPT \cite{men2024shortgpt} and LaCo \cite{yang2024laco}, or other structural units like attention heads and neuron columns \cite{michel2019sixteenheads, ma2023llmpruner, ashkboos2024slicegpt}.

Our contributions are threefold:
\begin{enumerate}
    \item We propose a two-component compression framework that combines an SVD-compressed primary path with a parallel, learnable, low-rank corrective path to restore functional performance.
    \item We introduce a data-informed, efficient procedure for selectively targeting specific LLM layers for compression based on their functional stability, followed by a joint fine-tuning process to minimize output error.
    \item Through empirical validation on three decoder-only LLMs, we demonstrate that CALR achieves superior compression-performance trade-offs compared to leading low-rank and structured pruning methods, particularly under aggressive compression.
\end{enumerate}

CALR advances LLM compression by explicitly modeling and correcting for functional information loss. This enables the creation of smaller, more efficient models without substantial performance degradation, expanding their practical deployment possibilities.

\section{Motivation for CALR}
\label{sec:motivation_related_work}

The CALR design is motivated by the intrinsic limitation of SVD compression and an empirical analysis of information processing dynamics within LLMs. Essentially, SVD allows an optimal low-rank approximation in terms of $L_2$ (Frobenius) norm minimization; however, information lost in the process may include important but functionally integral aspects of an LLM's function. Aggressive SVD truncation often leads to a noticeable decline in downstream task performance.

This observation forms the main hypothesis of CALR: if a primary low-rank approximation, via Singular Value Decomposition (SVD), is able to capture the dominant information in a layer, then one can construct a secondary mechanism that is parameter-efficient and can learn to recover the functionally important information lost during this approximation. It is hypothesized that the residual difference between the output of the original module, $F_{orig}(\mathbf{X})$, and the output of the SVD-compressed module, $F_{SVD}(\mathbf{X})$, contains this crucial corrective information. The additive, learnable, low-rank module of CALR is explicitly designed to model this functional residual.

In this study, we analyzed the heterogeneous nature of the mechanisms behind the information process in different large language models (LLMs) in order to empirically justify the imperative for an adaptable corrective mechanism. We analyzed layer-wise dynamics in six diverse open-source models: SmolLM2-135M, SmolLM2-360M \cite{allal2025smollm2}, Qwen3-0.6B, Qwen3-1.7B \cite{yang2025qwen3}, Llama-3.2-1B \cite{meta2024llama1b}, and Llama-3.2-3B \cite{meta2024llama3b}. We measured the transformation within each transformer block (Layer Block) and its Multilayer Perceptron (MLP) sub-block using the cosine distance between the block's input $\mathbf{X}_{in}$ and output $\mathbf{X}_{out}$:

\begin{equation}
    \text{Transformation}(\mathbf{X}_{in}, \mathbf{X}_{out}) = 1 - \frac{\mathbf{X}_{in} \cdot \mathbf{X}_{out}}{\|\mathbf{X}_{in}\|_2 \ \|\mathbf{X}_{out}\|_2}
    \label{eq:transformation_metric}
\end{equation}

\begin{figure}[t!]
  \centering
  \includegraphics[width=\columnwidth]{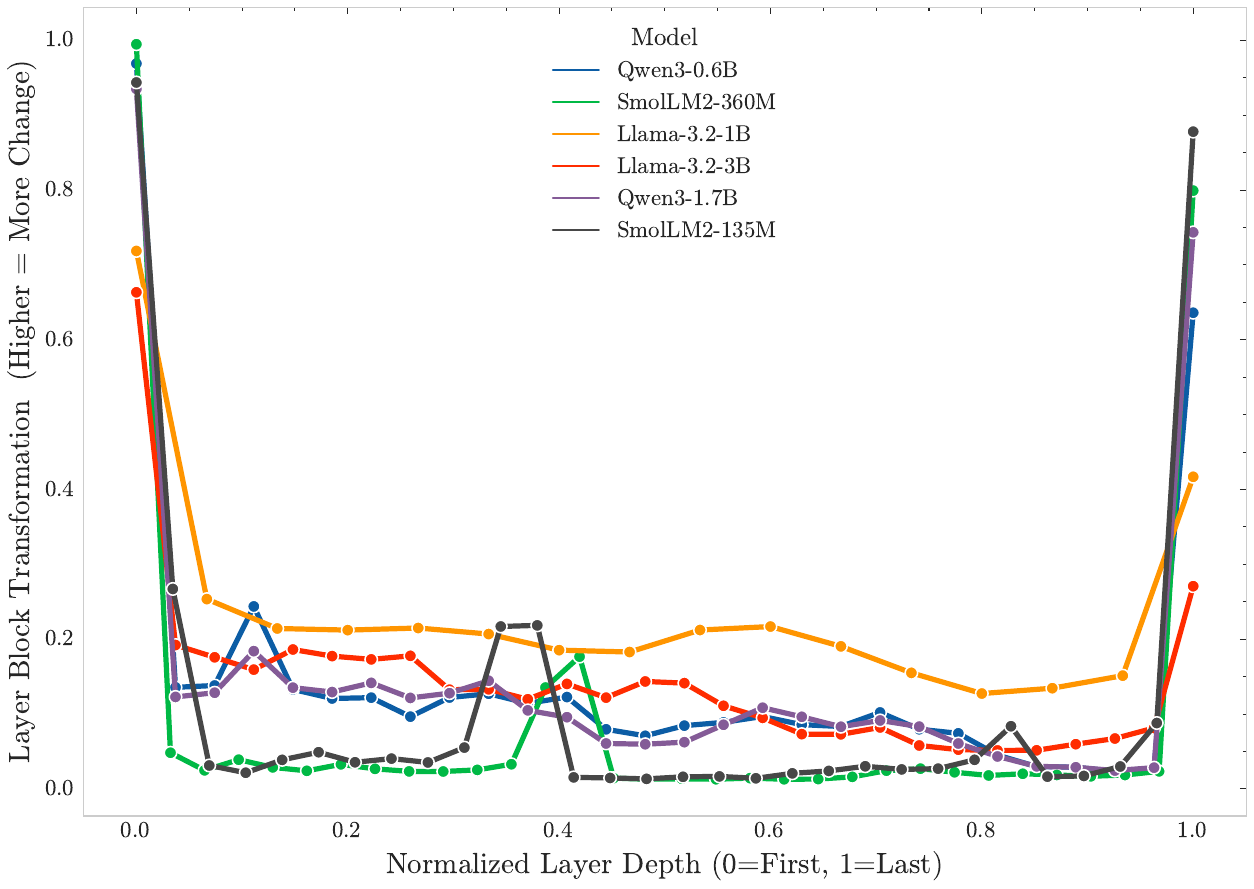}
  \caption{Layer Block Transformation across normalized layer depth for various LLMs. 
  % The characteristic "U-shape" observed in many models indicates that initial and final layers tend to perform more substantial representational changes compared to middle layers.
  }
  \label{fig:layer_block_transformation}
\end{figure}

As shown in Fig. \ref{fig:layer_block_transformation}, the overall Layer Block Transformation frequently exhibits a ``U-shape" pattern, indicating that initial and final layers perform more significant representational changes. This aligns with findings on ``cornerstone layers" \cite{zhang2024layerimportance} and suggests these layers are particularly sensitive to information loss \cite{rajapaksha2024}. 
This diverse, model-dependent behavior demonstrates that a uniform compression approach is suboptimal. 
CALR's learnable corrective module is specifically designed to address this challenge by adaptively learning to offset the functional limitations of the primary SVD-compressed module, offering a more robust solution for high-fidelity model compression.
% Furthermore, the MLP Block Transformation (Fig. \ref{fig:mlp_block_transformation}) reveals additional heterogeneity, with the correlation between layer and MLP transformation varying significantly across models (Fig. \ref{fig:layer_mlp_correlation}). 
% Architectural choices likely influence this behavior, and this also align with other finding. He et al. \cite{he2024matterstransformersattentionneeded} quantitatively evaluate redundancy across Transformers and observe that many attention layers exhibit high input–output similarity that indicate redundancy, whereas MLP layers are substantially less redundant. Altogether, this literature supports our interpretation of the Pearson correlation metric: high MLP-layer alignment signifies that the MLP is doing most of the model’s representational work, validating compression strategies that focus on MLP sub-blocks in such cases.

% \begin{figure}[t!]
%   \centering
%   \includegraphics[width=\columnwidth]{fig/mlp_block_importance_plot.pdf}
%   \caption{MLP Block Transformation across normalized layer depth for various LLMs. 
%   % The degree of transformation within MLP blocks also varies with depth, often showing an increasing trend, particularly in larger models. This highlights heterogeneity even within specific sub-components of layers.
%   }
%   \label{fig:mlp_block_transformation}
% \end{figure}

% This diverse, model-dependent behavior demonstrates that a uniform compression approach is suboptimal. 
% % Even with layer-specific rank selection, standard SVD may fail to preserve functionally important residual information differently across layers and models. 
% CALR's learnable corrective module is specifically designed to address this challenge by adaptively learning to offset the functional limitations of the primary SVD-compressed module, offering a more robust solution for high-fidelity model compression.

% \begin{figure}[t!]
%   \centering
%   \includegraphics[width=\columnwidth]{fig/layer_mlp_correlation_barplot.pdf}
%   \caption{Pearson correlation coefficient between Layer Block Transformation and MLP Block Transformation across different models. 
%   % The wide variation in correlation (from strongly negative to strongly positive) underscores significant model-specific differences in how MLP transformations contribute to overall layer transformations.
%   }
%   \label{fig:layer_mlp_correlation}
% \end{figure}

\section{Proposed Method}
\label{sec:proposed-method}

CALR achieves compression through a two-stage architectural transformation applied to targeted Feedforward Network (FFN) blocks, which are major contributors to an LLM's parameter count, as seen in Fig. \ref{fig:calr_architecture}. The process involves:

\begin{enumerate}
    \item Primary SVD-Based Compression: Each constituent linear layer within a target FFN module is replaced by its low-rank approximation derived from SVD.
    \item Additive Low-Rank Corrective Module: A separate, learnable, low-rank module is introduced. It operates in parallel with the SVD-compressed module, taking the same input, and its output is additively combined to produce the final, corrected output.
\end{enumerate}

\subsection{Architectural Transformation with CALR}
\label{ssec:architectural_transformation_calr}

The CALR framework transforms a target FFN module by first factorizing its primary linear layers and then introducing a parallel corrective path to restore function.

\subsubsection{Primary SVD-Based Compression}
\label{sssec:primary_svd_compression}

A standard gated FFN in an LLM accepts an input $\mathbf{X} \in \mathbb{R}^{b \times d_{\text{model}}}$ and computes its output $F_{\text{orig}}(\mathbf{X})$ as:
\begin{equation}
F_{\text{orig}}(\mathbf{X}) = \left[ (\mathbf{X} \mathbf{W}_g) \odot \text{ACT}(\mathbf{X} \mathbf{W}_u) \right] \mathbf{W}_d,
\label{eq:ffn_original}
\end{equation}
where $\mathbf{W}_g, \mathbf{W}_u, \mathbf{W}_d$ are the gate, up, and down projection weight matrices, $\odot$ is element-wise multiplication, and $\text{ACT}(\cdot)$ is a non-linear activation (e.g., SiLU).

CALR compresses this by replacing each weight matrix $\mathbf{W}_j$ (for $j \in \{g, u, d\}$) with a low-rank approximation $\mathbf{W}_j \approx \mathbf{A}_j \mathbf{B}_j$, initialized via truncated SVD. For a given matrix $\mathbf{W}_j = \mathbf{U}_j \mathbf{\Sigma}_j \mathbf{V}_j^\intercal$, its rank-$r$ factors are initialized as:
\begin{align}
    \mathbf{A}_j &\leftarrow \mathbf{U}_{j, :, :r} \mathbf{\Sigma}_{j, :r, :r}^{1/2}, \label{eq:calr_A_init} \\
    \mathbf{B}_j &\leftarrow \mathbf{\Sigma}_{j, :r, :r}^{1/2} \mathbf{V}_{j, :, :r}^\intercal, \label{eq:calr_B_init}
\end{align}
where $\mathbf{U}_{j, :, :r}$, $\mathbf{V}_{j, :, :r}^\intercal$, and $\mathbf{\Sigma}_{j, :r, :r}$ are the top $r$ singular vectors and values. This yields the best rank-$r$ approximation in Frobenius norm~\cite{Eckart_Young_1936}. The compressed FFN function, $F_{\text{SVD}}(\mathbf{X})$, becomes:
\begin{equation}
F_{\text{SVD}}(\mathbf{X}) = \left[ (\mathbf{X} \mathbf{A}_g \mathbf{B}_g) \odot \text{ACT}(\mathbf{X} \mathbf{A}_u \mathbf{B}_u) \right] \mathbf{A}_d \mathbf{B}_d.
\label{eq:ffn_svd}
\end{equation}
These factor matrices $\mathbf{A}_j$ and $\mathbf{B}_j$ constitute the primary computation path and are trainable.

\subsubsection{The Additive Low-Rank Corrective Module}
\label{sssec:corrective_module_calr}

CALR introduces a parallel corrective module, $F_{\text{corr}}(\mathbf{X})$, defined by two trainable low-rank matrices $\mathbold{\mathcal{A}} \in \mathbb{R}^{d_{\text{model}} \times r_c}$ and $\mathbold{\mathcal{B}} \in \mathbb{R}^{r_c \times d_{\text{model}}}$, where $r_c$ or $\operatorname{rank}(\mathbold{\mathcal{A}} \  \mathbold{\mathcal{B}}) = r_c$ is the corrective rank (typically $r_c=r$). The computation is:
\begin{equation}
F_{\text{corr}}(\mathbf{X}) = X(\mathbold{\mathcal{A}} \  \mathbold{\mathcal{B}}),
\end{equation}
The final output of the CALR-transformed FFN is the sum of the primary and corrective paths:
\begin{align}
F_{\text{CALR}}(\mathbf{X}) &= F_{\text{SVD}}(\mathbf{X}) + F_{\text{corr}}(\mathbf{X}) \nonumber \\
&= F_{\text{SVD}}(\mathbf{X}) + X(\mathbold{\mathcal{A}} \  \mathbold{\mathcal{B}}).
\label{eq:calr_final_output}
\end{align}
This enables the module to recover the ``functional residual" lost during SVD compression.

\subsubsection{Formalizing the Corrective Objective: Functional vs. Weight Residual}
\label{sssec:formalizing_objective}

The CALR design is motivated by the goal of correcting errors in the function output space, which is more directly aligned with preserving task performance than correcting errors in the weight space. The functional residual $R_F(X)$ is defined as:
\begin{equation}
R_F(\mathbf{X}) = F_{\text{orig}}(\mathbf{X}) - F_{\text{SVD}}(\mathbf{X}).
\end{equation}
The corrective module aims to approximate $R_F(X)$ by minimizing the functional objective:
\begin{equation}
\min_{\mathbold{\mathcal{A}}, \mathbold{\mathcal{B}}} \; \mathbb{E}_{\mathbf{X} \sim \mathcal{D}} 
\left\| F_{\text{orig}}(\mathbf{X}) 
- \left( F_{\text{SVD}}(\mathbf{X}) + X(\mathbold{\mathcal{A}} \  \mathbold{\mathcal{B}}) \right) \right\|^2,
\label{eq:calr_functional_objective}
\end{equation}
where $\mathcal{D}$ is the data distribution. Unlike weight-based approaches that minimize $\|\mathbf{W} - \mathbf{AB}\|_F$, this explicitly targets functional fidelity.

\subsubsection{Hypothesis and Empirical Justification}
\label{sssec:hypothesis_justification}

The effectiveness of CALR relies on the following foundational hypothesis: the functional residual, $R_F(\mathbf{X}) = F_{\text{orig}}(\mathbf{X}) - F_{\text{SVD}}(\mathbf{X})$, reflects the information lost during the SVD compression process and is marked by a low-rank structure. While $F_{\text{orig}}$ is highly non-linear, we assume that the \emph{discrepancy} between this function and its high-fidelity SVD counterpart represents a more tractable and structured function that can be approximated with more effectiveness.

\begin{figure}[t!]
  \centering
  \includegraphics[width=\columnwidth]{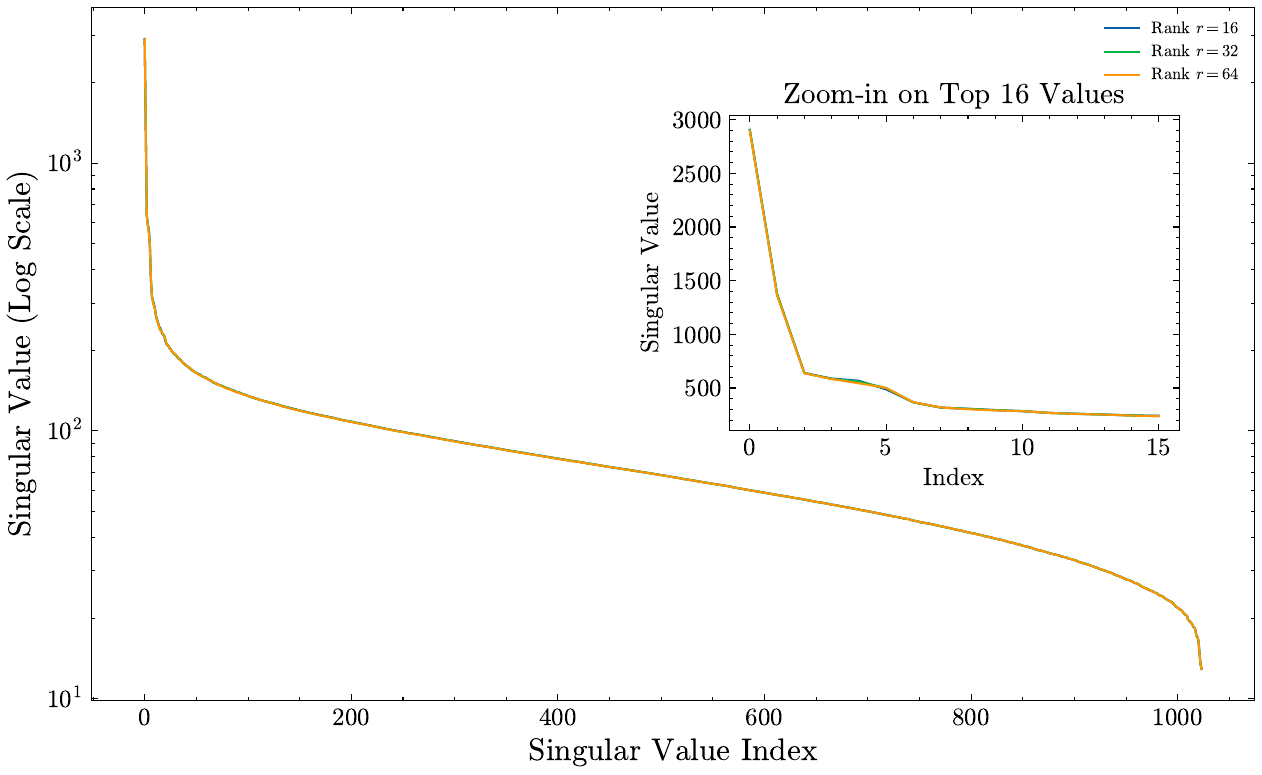}
  \caption{Singular value spectrum of the functional residual at the layer with the lowest transformation score, showing rapid spectral decay and low-rank structure. Inset highlights top-16 singular values across different SVD ranks.}
  \label{fig:residual_spectrum}
\end{figure}

To provide \emph{a priori} evidence for this hypothesis, we performed an analysis of the functional residual for a low-transformation layer in the Qwen3-0.6B model. The residual matrix was calculated using a calibration data set, and then its singular value spectrum was analyzed as shown in Fig. \ref{fig:residual_spectrum}.

The results are compelling. The spectral analysis demonstrates a significant drop, which itself is a strong indicator of a low-rank matrix and, in turn, justifies that the missing data shows a coherent pattern and not random noise. The inset demonstrates that the scale of this residual has an inverse relationship with the quality of the original singular value decomposition (SVD) approximation. This is a strong justification of our choice of design. Since the error signal itself is inherently of low-rank nature, an adaptive low-rank corrective module $F_{\text{corr}}(\mathbf{X}) = \mathbold{\mathcal{A}} \ \mathbold{\mathcal{B}}$ stands out as a parameter-sparse and optimal choice for its modeling and correction. In conclusion, the error of the CALR approximation is measured as the norm of the amount omitted in this residual  $\| R_F(\mathbf{X}) - F_{\text{corr}}(\mathbf{X}) \|$. The effectiveness of our method, demonstrated in Section \ref{sec:result_analysis}, also justifies this underlying principle.

\subsection{Training and Optimization}
\label{ssec:training_optimization_calr}

The CALR framework uses a two-phase optimization process detailed in Algorithm \ref{alg:calr_construction_training}.

\begin{algorithm}[t]
\caption{CALR Construction and Fine-Tuning}
\label{alg:calr_construction_training}
\begin{algorithmic}[1]
\STATE Input: Pre-trained model $\mathcal{M}_{orig}$, number of target FFNs $N_{target}$, SVD rank $r$, corrective rank $r_c$ (default $r_c = r$), calibration data $\mathcal{D}_{cal}$, training hyperparameters.
\STATE Set working model: $\mathcal{M}_{CALR} \leftarrow \mathcal{M}_{orig}$.
\STATE \textbf{Phase 1: Module Selection and Decomposition}
\STATE Compute transformation metric (as defined in \ref{eq:transformation_metric}) for each FFN using $\mathcal{D}_{cal}$.
\STATE Select $N_{target}$ FFNs with the lowest transformation values $\rightarrow \mathcal{F}_{target}$.
\FOR{each module $m \in \mathcal{F}_{target}$}
    \FOR{each linear layer $\mathbf{W}_j$ in $m$}
        \STATE Apply SVD: $\mathbf{W}_j = \mathbf{U}_j \mathbf{\Sigma}_j \mathbf{V}_j^\top$.
        \STATE Replace $\mathbf{W}_j \leftarrow \mathbf{A}_j \mathbf{B}_j$, where:\par
        \hskip\algorithmicindent $\mathbf{A}_j = \mathbf{U}_{j,:,:r} \mathbf{\Sigma}_{j,:r,:r}^{1/2}$,\quad $\mathbf{B}_j = \mathbf{\Sigma}_{j,:r,:r}^{1/2} \mathbf{V}_{j,:,:r}^\top$.
    \ENDFOR
    \STATE Initialize corrective module: $\mathbold{\mathcal{A}}_m$ (e.g., Kaiming), $\mathbold{\mathcal{B}}_m$ (e.g., zeros).
    \STATE Add parallel path: $\mathbold{\mathcal{A}}_m \mathbold{\mathcal{B}}_m$ to $m$ as per (\ref{eq:calr_final_output}).
\ENDFOR
\STATE \textbf{Phase 2: Fine-Tuning}
\STATE Define trainable params: all $\{\textbf{A}_j, \textbf{B}_j, \mathbold{\mathcal{A}}_m, \mathbold{\mathcal{B}}_m\}$ for $m \in \mathcal{F}_{target}$.
\STATE Fine-tune $\mathcal{M}_{CALR}$ on $\mathcal{D}_{cal}$ using a language modeling loss, $\mathcal{L}_{LM}$.
\STATE Return: Fine-tuned, compressed model $\mathcal{M}_{CALR}$.
\end{algorithmic}
\end{algorithm}

\subsubsection{Selective Initialization and Transformation.}
First, we specify the number of FFN modules to compress, $N_{target}$. We then calculate the transformation value (using  (\ref{eq:transformation_metric})) for all FFNs on a small calibration dataset. We select the $N_{target}$ modules with the \textit{lowest} transformation values, as these more stable layers are likely more amenable to SVD compression. For each linear layer within these modules, we initialize its low-rank factors $\mathbf{A}_j, \mathbf{B}_j$ using SVD (as defined in (\ref{eq:calr_A_init}) and  (\ref{eq:calr_B_init})). We also initialize a corresponding corrective module, setting its $\mathbold{\mathcal{B}}$ matrix to zeros (inspired by LoRA \cite{hu2021lora}) so it initially has no effect.

\subsubsection{Joint Fine-tuning.}
After initialization, all new low-rank parameters (both primary factors $\mathbf{A}_j, \mathbf{B}_j$ and corrective factors $\mathbold{\mathcal{A}}, \mathbold{\mathcal{B}}$) are made trainable. The entire compressed model is then fine-tuned on a calibration dataset using a standard language modeling objective. This joint optimization allows the primary and corrective components to co-adapt. The SVD-initialized factors can deviate from their initial values to better align with the overall task objective, while the corrective module learns to produce an output that minimizes the functional difference between the modified module and its uncompressed original, thereby recovering lost performance.

\section{Experiments}
\label{sec:experiments}

This section details the experimental evaluation of CALR, following the scientific method to test our hypothesis. We outline the setup, baselines, and evaluation protocols designed to ensure experimental validity.

\subsection{Experimental Setup}
\label{ssec:experimental_setup}

\subsubsection{Base Models and CALR Configuration}
\label{sssec:model_config_calr}
Experiments are conducted on three decoder-only LLMs of varying sizes: SmolLM2-135M \cite{allal2025smollm2}, Qwen3-0.6B \cite{yang2025qwen3}, and Llama-3.2-1B \cite{meta2024llama1b}. For CALR, we set the corrective rank equal to the primary rank ($r_c = r$). For our main experiments, we selected a rank of $r=32$. This value was empirically determined to offer an optimal trade-off between performance recovery and parameter efficiency, as detailed in our sensitivity analysis in Section \ref{sssec:ablation_corrective_rank}. The number of targeted FFNs ($N_{\text{target}}$) for each model was then chosen to meet specific compression goals, as shown in Table \ref{tab:calr_configuration}.

\begin{table}[h!]
\caption{CALR Configuration}
\centering
\renewcommand{\arraystretch}{1.3}
\begin{tabular}{p{2cm} |  c  c c }
\hline
\textbf{Base Model} & $N_{target}$ & $r$ & $r_c$ \\
\hline
\multirow{2}{*}{\textbf{SmolLM2-135M}} 
    & 4      & 32 &  32 \\ 
    & 13     & 32 &  32 \\
\hline
\multirow{2}{*}{\textbf{Qwen3-0.6B}} 
    & 8      & 32 &  32 \\ 
    & 16     & 32 &  32 \\
\hline
\multirow{2}{*}{\textbf{Llama-3.2-1B}} 
    & 6      & 32 &  32 \\ 
    & 11     & 32 &  32 \\

\hline
\end{tabular}
\renewcommand{\arraystretch}{1}  % Reset to default value if needed
\label{tab:calr_configuration}
\end{table}

\subsubsection{Baselines}
\label{sssec:baselines_calr}
To ensure a robust comparison, we evaluate CALR against the top competitive algorithms representing distinct, mainstream compression strategies. The choice of these baselines is justified in detail in our discussion (Section \ref{ssec:limitations_and_scope}).
\begin{itemize}
    \item Base Model: The original, unmodified pre-trained model serves as the performance upper bound.
    \item LaCo (Layer Collapse) \cite{yang2024laco}: A leading layer-removal method that merges layers, representing the ``depth reduction" paradigm.
    \item ShortGPT (Layer Pruning) \cite{men2024shortgpt}: Another state-of-the-art layer-removal method that prunes entire layers based on an importance score.
    \item LoSparse (Low-Rank and Sparse Approximation) \cite{li2023losparse}: Our most direct conceptual competitor. It uses a low-rank plus \textit{sparse} corrective term, providing a clear test of our hypothesis that a \textit{dense} corrective term is superior for this context.
\end{itemize}

\subsubsection{Calibration and Initialization}
\label{sssec:calibration_calr}
The selective compression strategy is guided by calculating the layer transformation metric (as defined in (\ref{eq:transformation_metric})) on 100 random samples from the Cosmopedia dataset \cite{le2024cosmopedia}. This data-informed process for selecting ``quieter" layers is computationally inexpensive. CALR's SVD and corrective factors are initialized as described in Section \ref{ssec:training_optimization_calr}.

\subsubsection{Compression Targets}
\label{sssec:compression_targets_calr}
We evaluate two compression configurations for each model:
\begin{enumerate}
    \item LaCo-aligned: Compression is set to match the parameter reduction of LaCo, ensuring a direct comparison point.
    \item Aggressive Compression: Compression is set to at least a 25\% parameter reduction to test robustness.
\end{enumerate}
Parameters for ShortGPT and LoSparse are adjusted to meet these same targets.

\subsubsection{Supervised Fine-Tuning (SFT)}
\label{sssec:sft_protocol_calr}
To recover performance, all compressed models (CALR and baselines) undergo an identical SFT procedure. The base models are \textit{not} fine-tuned.
\begin{itemize}
    \item Training Steps: 5120 steps.
    \item {Batch Size}: Effective batch size of 16.
    \item {Precision}: fp32 for smaller models, fp16 for Llama-3.2-1B on a single NVIDIA V100 32GB GPU.
    \item {Dataset Mixture}: A combination of Cosmopedia \cite{le2024cosmopedia}, MMLU \cite{hendrycks2021measuring}, HellaSwag \cite{zellers2019hellaswag}, ARC \cite{clark2018think}, OpenBookQA \cite{mihaylov2018can}, and GLUE \cite{wang2018glue}.
    \item {Optimizer}: AdamW \cite{loshchilov2017decoupled} with a cosine learning rate schedule (LR=$2e-5$, weight decay=0.01).
    \item {Trainable Parameters}: All parameters of the compressed models are fine-tuned.
\end{itemize}

\subsubsection{Evaluation}
\label{sssec:evaluation_protocol_calr}

To ensure fair and rigorous evaluation, the SFT data is strictly separated from the final evaluation data. We evaluate on the non-overlapping validation/test splits of a comprehensive suite of English natural language understanding (NLU) tasks:
\begin{itemize}
    \item ARC-Easy/Challenge (5-shot) \cite{clark2018think}
    \item HellaSwag (5-shot) \cite{zellers2019hellaswag}
    \item MMLU (0-shot) \cite{hendrycks2021measuring}
    \item OpenBookQA (5-shot) \cite{mihaylov2018can}
    \item GLUE Benchmark Subset (0-shot) \cite{wang2018glue} (MNLI, QNLI, RTE, SST-2, WNLI).
\end{itemize}
All evaluations are conducted using the EleutherAI Language Model Evaluation Harness \cite{gao2021framework} for standardization and reproducibility.

\subsubsection{Implementation Details}
\label{sssec:implementation_details_calr}
Experiments use PyTorch and the Hugging Face Transformers library \cite{wolf2020transformers}. Consistent hyperparameters are used across all methods for fairness. Code is available at our repository.
\footnote{\href{https://github.com/xxxx/calr.git}{\underline{\textit{Github: https://github.com/xxxx/calr.git}}}}

\section{Result and Analysis}
\label{sec:result_analysis}

\subsection{Performance Comparison with Baselines}
\label{ssec:performance_comparison_calr}

\begin{table*}[h!]
\caption{Comparison of CALR with base models and compression baselines on SmolLM2-135M, evaluated across multiple benchmarks and parameter reduction targets. }
\centering
\renewcommand{\arraystretch}{1.3}
\begin{tabular}{l c c | c c c c c c | c}
\hline
\multirow{2}{*}{\textbf{Method}}  &  \multirow{2}{*}{\textbf{Reduction}} & \multirow{2}{*}{\textbf{Params (M)}} & \multicolumn{7}{c}{\textbf{Benchmark}} \\
\cline{4-10}
& & & \textbf{ARC-C} & \textbf{ARC-E} & \textbf{HellaSwag} & \textbf{MMLU} & \textbf{GLUE} & \textbf{OBQA} & \textbf{Average} \\
\hline
Base      & 0\%      & 134.52 & 29.61\% & 58.59\% & 43.12\% & 24.36\% & 49.59\% & 21.80\% & 37.48\% \\
\hline
\multicolumn{9}{c}{\textit{Moderate Compression Target ($\sim$10\%)}} \\
\hline
LaCo      & 10.53\%  & 120.35 & 25.94\% & 39.81\% & 34.15\% & \textbf{24.27\%} & 46.09\% & 18.80\%  & 31.51\% \\
ShortGPT  & 10.53\%  & 120.35 & 26.28\% & 49.12\% & 36.24\% & 23.02\% & \textbf{48.27\%} & 18.20\% & 33.52\%  \\
LoSparse  & 10.19\%  & 120.80 & 25.34\% & 24.66\% & 26.44\% & 22.95\% & 46.96\% & 18.40\% & 27.46\%  \\
CALR      & \textbf{10.77\%} & \textbf{120.03} & \textbf{27.39\%} & \textbf{49.54\%} & \textbf{36.72\%} & 23.78\% & 46.10\% & \textbf{19.80\%} & \textbf{33.89\%}  \\
\hline

\multicolumn{9}{c}{\textit{Aggressive Compression Target ($\sim$26\%)}} \\

\hline
ShortGPT  & 26.32\%  & 99.11  & 23.55\% & 26.14\% & 27.26\% & 23.28\% & 46.15\% & 16.00\% & 27.06\%  \\
LoSparse  & 26.74\%  & 98.55  & \textbf{25.00\%} & 25.63\% & 25.15\% & \textbf{24.66\%} & 46.03\% & \textbf{19.40\%} & 27.65\%  \\
CALR      & \textbf{26.93\%} & \textbf{98.30} & 23.12\% & \textbf{33.54\%} & \textbf{27.83\%} & 23.14\% & \textbf{46.86\%} & 15.00\% & \textbf{28.25\% } \\
\hline
\end{tabular}
\renewcommand{\arraystretch}{1}  % Reset to default 
\label{tab:smollm_ieee}

\end{table*}

\begin{table*}[h!]
\caption{Comparison of CALR with base models and compression baselines on Qwen3 0.6B, evaluated across multiple benchmarks and parameter reduction targets. }
\centering
\renewcommand{\arraystretch}{1.3}
\begin{tabular}{l c c | c c c c c c | c}
\hline
\multirow{2}{*}{\textbf{Method}}  &  \multirow{2}{*}{\textbf{Reduction}} & \multirow{2}{*}{\textbf{Params (M)}} & \multicolumn{7}{c}{\textbf{Benchmark}} \\
\cline{4-10}
& & & \textbf{ARC-C} & \textbf{ARC-E} & \textbf{HellaSwag} & \textbf{MMLU} & \textbf{GLUE} & \textbf{OBQA} & \textbf{Average} \\
\hline
Base      & 0\%     & 596.05 & 34.13\% & 55.98\% & 47.30\% & 40.17\% & 49.29\% & 21.00\% &  41.31\%\\
\hline

\multicolumn{9}{c}{\textit{Moderate Compression Target ($\sim$14\%)}} \\
\hline

LaCo      & 13.20\% & 517.40 & \textbf{27.82\%} & 38.05\% & 34.65\% & 24.01\% & 46.16\% & 15.60\% &  31.05\% \\
ShortGPT  & 13.20\% & 517.40 & 26.45\% & 35.44\% & 34.02\% & 23.05\% & 44.99\% & \textbf{17.80\%} &  30.29\% \\
LoSparse  & 14.74\% & 508.20 & 26.19\% & 24.45\% & 26.36\% & \textbf{24.62\%} & 44.94\% & 15.20\% &  26.96\% \\
CALR      & \textbf{15.45\%} & \textbf{503.97} & 27.13\% & \textbf{45.20\%} & \textbf{37.72\%} & 23.36 & \textbf{50.79\%} & 16.80\% &  \textbf{33.50\%}\\
\hline
\multicolumn{9}{c}{\textit{Aggressive Compression Target ($\sim$27\%)}} \\
\hline
ShortGPT  & 26.39\% & 438.74 & 26.71\% & 36.15\% & 33.42\% & 24.02\% & 46.81\% & 17.60\% &  30.79\%\\
LoSparse  & 27.26\% & 433.54 & 24.57\% & 23.65\% & 20.51\% & \textbf{24.37\%} & 49.27\% & \textbf{18.20\%} &  26.76\%\\
CALR      & \textbf{27.81\%} & \textbf{430.31} & \textbf{26.88\%} & \textbf{44.99\%} & \textbf{37.41\%} & 23.25\% & \textbf{49.97\%} & 16.80\% &  \textbf{33.22\%} \\
\hline
\end{tabular}
\renewcommand{\arraystretch}{1}  % Reset to default 
\label{tab:qwen}
\end{table*}

\begin{table*}[h!]
\caption{Comparison of CALR with base models and compression baselines on Llama-3.2-1B, evaluated across multiple benchmarks and parameter reduction targets. }
\centering
\renewcommand{\arraystretch}{1.3}
\begin{tabular}{l c c | c c c c c c | c}
\hline
\multirow{2}{*}{\textbf{Method}}  &  \multirow{2}{*}{\textbf{Reduction}} & \multirow{2}{*}{\textbf{Params (M)}} & \multicolumn{7}{c}{\textbf{Benchmark}} \\
\cline{4-10}
& & & \textbf{ARC-C} & \textbf{ARC-E} & \textbf{HellaSwag} & \textbf{MMLU} & \textbf{GLUE} & \textbf{OBQA} & \textbf{Average} \\
\hline
Base      & 0\%     & 1235.81 & 36.26\% & 60.48\% & 63.66\% & 37.69\% & 51.95\% & 26.40\% & 46.07\% \\
\hline
\multicolumn{9}{c}{\textit{Moderate Compression Target ($\sim$30\%)}} \\
\hline
LaCo      & 29.53\% & 870.89  & 27.99\% & 30.09\% & 37.32\% & \textbf{24.38\%} & 46.42\% & 16.00\% & 30.37\% \\
ShortGPT  & 29.53\% & 870.89  & \textbf{29.27\%} & 30.30\% & \textbf{37.55\%} & 23.78\% & 45.81\% & 14.80\% & 30.25\% \\
LoSparse  & 29.87\% & 866.67  & 27.82\% & 27.06\% & 26.06\% & 24.28\% & 45.49\% & 15.20\% & 27.65\% \\
CALR      & \textbf{31.86\%} & \textbf{842.07} & 25.51\% & \textbf{36.87\%} & 35.58\% & 23.54\% & \textbf{48.46\%} & \textbf{17.80\%} & \textbf{31.29\%} \\

\hline
\multicolumn{9}{c}{\textit{Aggressive Compression Target ($\sim$50\%)}} \\
\hline

ShortGPT  & 49.22\% & 627.60  & 24.23\% & 28.03\% & \textbf{27.74\%} & \textbf{25.17\%} & 44.82\% & 15.20\% & \textbf{27.53\%} \\
LoSparse  & 49.58\% & 623.16  & \textbf{26.02\%} & 26.22\% & 25.07\% & 24.61\% & 44.40\% & \textbf{16.80\%} & 27.19\% \\
CALR      & \textbf{51.77\%} & \textbf{595.99} & 24.23\% & \textbf{29.50\%} & 27.29\% & 23.47\% & \textbf{45.03\%} & 14.80\% & 27.39\% \\
\hline
\end{tabular}
\renewcommand{\arraystretch}{1}  % Reset to default 
\label{tab:llama}
\end{table*}

An exhaustive evaluation of CALR compared with baseline models and three standard compression benchmarks on two parameter reduction tasks. The results are shown in Tables \ref{tab:smollm_ieee}, \ref{tab:qwen}, and \ref{tab:llama}.

In the three models and for moderate and aggressive compression targets, CALR consistently displays a strong balance between compression efficacy and preservation of performance. Notably, CALR often achieves the greatest parameter reduction within the given range under each set compression target while delivering competitive or superior results on the evaluated benchmarks.

For the SmolLM2-135M model, as shown in Table \ref{tab:smollm_ieee}, at 10.77\% reduction, CALR surpasses other methods on ARC-C, ARC-E, HellaSwag, and OpenBookQA, retaining approximately 90.42\% of the base model's average performance. At a more aggressive 26.93\% reduction, CALR continues to lead on several key benchmarks, outperforming ShortGPT and LoSparse in the overall average performance.

A similar trend is observed in the Qwen3-0.6B model (Table \ref{tab:qwen}). At both 15.45\% and 27.81\% reduction targets, CALR shows marked improvements, particularly in those tasks with extensive reasoning needs like ARC-E and general language understanding (GLUE). In both scenarios, CALR not only achieves the greatest reduction in parameters but also shows better overall average performance compared to other compressed models.

Compared to the larger Llama-3.2-1B model in Table  \ref{tab:llama}, CALR demonstrates its scalability nature. At a substantial 31.86\% parameter reduction, CALR leads on ARC-E, GLUE, and OpenBookQA. Even at a very high reduction target of 51.77\%, CALR remains highly competitive, particularly on ARC-E and GLUE, while offering the most significant parameter savings and the smallest final model footprint.

The results demonstrate the effectiveness of CALR's two-component approach. The SVD-based method supports strong compression, while the corrective module, which is defined by tunable parameters, is essential to minimize information loss and restore functional capabilities.

\subsection{Statistical Significance Testing}
\label{ssec:statistical_testing_calr}

To ensure the scientific integrity of our claims, we used non-parametric statistical tests ($\alpha = 0.05$). We first used the Friedman test \cite{Friedman01121937} to determine general differences between groups, and then pairwise Wilcoxon signed-rank tests \cite{wilcoxon1945}, using the Holm-Bonferroni correction \cite{holm1979simple}. The hypotheses for each of these tests were defined as follows:

\begin{itemize}
    \item Null hypothesis ($H_0$): The median difference in performance between CALR and the baseline is zero.
    \item Alternative hypothesis ($H_1$): CALR performs significantly better than the baseline.
\end{itemize}

The Friedman test on the overall scores of all the models and tasks at the lower compression level gave a p-value of $0.00160$, showing significant differences between the approaches. The results of the pair tests, as seen in Table \ref{tab:wilcoxon_overall_calr}, confirm that CALR shows a statistically significant lead over LoSparse and ShortGPT.

\begin{table}[h!]
\caption{Wilcoxon Signed-Rank Test Results (CALR vs. Baselines) with Holm-Bonferroni Correction ($\alpha = 0.05$). Significant p-values ($p < \alpha_{adj}$) are marked accordingly.}
\label{tab:wilcoxon_overall_calr}
\centering
\renewcommand{\arraystretch}{1.3}
\begin{tabular}{l | c c c}
\hline
\textbf{Comparison} & \textbf{p-value} & \textbf{Adjusted $\alpha$} & \textbf{Significant?} \\
\hline
CALR vs. LoSparse    & 0.00435       & 0.01667 ($0.05/3$)       & Yes \\
CALR vs. LaCo        & 0.02685       & 0.02500 ($0.05/2$)       & No  \\
CALR vs. ShortGPT    & 0.04394       & 0.05000 ($0.05/1$)       & Yes \\
\hline
\end{tabular}
\renewcommand{\arraystretch}{1}

\end{table}

Statistical analysis reveals that CALR achieves a significant performance improvement when set against the low-rank sparse approach (LoSparse) and the layer pruning method (ShortGPT). This affirms CALR's inherent competitiveness and its empirically proven dominance in retaining model performance after compression.

It is critical to note that the difference in overall performance between CALR and the layer-collapsing baseline LaCo did not meet the stringent statistical significance criterion ($p=0.02685 > \alpha_{adj}=0.02500$). However, the result needs to be read within the context of the compression-performance trade-off. As detailed in our primary results (Table ~\ref{tab:smollm_ieee}, \ref{tab:qwen}, and \ref{tab:llama}), CALR consistently achieves a greater parameter reduction than LaCo while delivering a higher average performance. For example, when compressing Llama-3.2-1B (Tables \ref{tab:llama}), CALR achieves a parameter reduction of 31.86\%, compared to the 29.53\% achieved by the reduction prompted by LaCo, yet achieves a higher average score (31.29\% vs. 30.37\%). This highlights that CALR uses a better parameter-efficient compression approach, enabling a better model size-operational performance trade-off. Thus, while LaCo is a good baseline model, the ability of CALR to achieve a smaller but higher-performing model makes it the better approach.

\subsection{Inference Efficiency and Latency}
\label{ssec:inference_efficiency}
While parameter count affects memory footprint, actual efficiency is determined by latency and throughput in the case of inference. Our compressed Llama-3.2-1B model, which obtained a remarkable compression ratio of about 30\%, was evaluated on a single NVIDIA V100 GPU as seen in Table \ref{tab:latency}.

\begin{table*}[h!]
\caption{Inference performance on Llama-3.2-1B at $\sim$29\% parameter reduction. 
% All benchmarks use a V100 GPU with fp16 precision and an enabled KV cache. Latency is measured per generated token. CALR demonstrates a strong balance of high throughput and superior accuracy.
}
\label{tab:latency}
\centering
\renewcommand{\arraystretch}{1.3}
\begin{tabular}{lccccc}
\hline
\textbf{Method} & \textbf{Reduction (\%)} & \textbf{Params (M)} & \textbf{Latency (ms/token)} & \textbf{Throughput (tok/s)} \\
\hline
Base Model        & 0.0\% & 1235.81  & 31.56 & 31.68 \\
\hline
% \multicolumn{5}{c}{\textit{Layer Removal Baselines}} \\
% \hline
LaCo              & 29.53\%  & 870.89 & \textit{19.92} & \textit{50.20} \\
ShortGPT          & 29.53\%  & 870.89 & \textbf{19.89} & \textbf{50.27} \\
% \hline
% \multicolumn{5}{c}{\textit{Low-Rank Baselines and CALR}} \\
\hline
LoSparse          & 29.87\%  & 866.67 & 50.22 & 19.91 \\
CALR-SVD$^{\mathrm{*}}$ & \textbf{32.03\%}  & \textbf{839.97} & 24.06 & 41.57 \\
\textbf{CALR} & \textit{31.86\%} & \textit{842.07} & 25.11 & 39.82 \\
\hline
\multicolumn{5}{l}{$^{\mathrm{*}}$CALR version without the Corrective Module, similar setup in Section \ref{sssec:ablation_corrective_module}}
\end{tabular}
\renewcommand{\arraystretch}{1}  % Reset to default 

\end{table*}

The results confirm CALR's practical efficiency. It achieves a 1.26x speedup over the base model. Critically, comparison with our SVD-Only ablation reveals that the corrective module adds only a negligible 4.40\% latency overhead, demonstrating that its substantial accuracy recovery comes at a minimal computational cost. Compared to its closest conceptual competitor, LoSparse, CALR is more than two times faster. While layer-removal methods (LaCo, ShortGPT) are fastest, they incur a greater performance penalty (Table \ref{tab:llama}). CALR offers a more compelling trade-off, delivering a significant speedup while retaining superior functional capabilities.

\subsection{Ablation Studies}
\label{ssec:ablation_studies_calr}
To understand the individual contributions of CALR's design choices, we conduct a series of ablation studies on the Qwen3-0.6B model.

\subsubsection{Impact of the Corrective Module ($F_{\text{corr}}$)}
\label{sssec:ablation_corrective_module}
To quantify the contribution of the corrective module, we compared three configurations:
\begin{itemize}
    \item Full CALR: Standard CALR with a fine-tuned primary path and corrective module.
    \item SVD-Only, Fine-tuned: The corrective module is removed, but the primary SVD factors are fine-tuned.
    \item SVD-Fixed, No Fine-tuning: The primary SVD factors are frozen post-initialization.
\end{itemize}

\begin{table}[h!]
\caption{Ablation on the corrective module ($F_{\text{corr}}$). 
% Results confirm the corrective module's critical role in recovering performance lost to SVD compression.
}
\centering
\renewcommand{\arraystretch}{1.3}
\begin{tabular}{p{2cm} |  c  c  c  }
\hline
\textbf{Benchmark} & \textbf{CALR} & \textbf{CALR SVD-Only} & \textbf{CALR SVD-Fixed} \\
\hline
\textbf{ARC-C} & \textbf{26.88\%} & 26.37\% & 26.45\% \\
\textbf{ARC-E} & \textbf{44.99\%} & 39.86\% & 27.40\% \\
\textbf{HellaSwag} & \textbf{37.41\%} & 34.74\% & 29.57\% \\
\textbf{MMLU} & 23.25\% & 22.92\% & \textbf{23.78\%} \\
\textbf{GLUE} & \textbf{47.94\%} & 45.27\% & 44.67\% \\
\textbf{OpenBookQA} & 16.80\% & \textbf{17.60\%} & 15.00\% \\
\hline
\textbf{Average} & \textbf{32.88\%} & 31.13\% & 27.81\% \\
\hline
\end{tabular}
\renewcommand{\arraystretch}{1}  % Reset to default value if needed
\label{tab:abl_a_result}

\end{table}

The results in Table \ref{tab:abl_a_result} unequivocally demonstrate the critical role of the corrective module. A clear performance hierarchy emerges: Full CALR $>$ SVD-Only $>$ SVD-Fixed.

On average, removing the corrective module (CALR SVD-Only) causes a 1.75 percentage point drop in performance. The improvement from CALR SVD-Only to Full CALR underscores that the dedicated corrective module is not redundant. Since it learns to compensate for residual functional errors that cannot be fully addressed by merely optimizing the primary SVD factors. This study validates that the corrective component is a vital element of CALR, essential for achieving superior performance.

\subsubsection{Sensitivity to Corrective Module Rank ($r_c$)}
\label{sssec:ablation_corrective_rank}

To determine an optimal rank for the corrective module, we analyzed its impact on performance by varying its rank, $r_c$, while holding the primary SVD rank constant at $r=32$. The results, presented in Table \ref{tab:abl_b_result}, demonstrate a clear, positive correlation between the corrective rank and average model performance, which improves from 30.82\% at $r_c=8$ to 34.12\% at $r_c=128$.

\begin{table}[h!]
\caption{Impact of corrective module rank ($r_c$) on CALR performance. 
% Increasing $r_c$ consistently improves average accuracy, highlighting the importance of corrective capacity in modeling the functional residual left by SVD.
}
\centering
\renewcommand{\arraystretch}{1.75}
\setlength{\tabcolsep}{5pt}
\begin{tabular}{p{1.6cm} | c c c c c }
\hline
\textbf{Benchmark} & $r_c=8$ & $r_c=16$ & $r_c=32$ & $r_c=64$ & $r_c=128$ \\
\hline
\textbf{ARC-C} & 26.11\% & 26.54\% & 26.88\% & \textbf{27.47\%} & \textbf{27.47\%} \\
\textbf{ARC-E} & 39.65\% & 39.65\% & 44.99\% & 45.20\% & \textbf{49.54\%} \\
\textbf{HellaSwag} & 35.50\% & 35.47\% & 37.41\% & 37.55\% & \textbf{37.72\%} \\
\textbf{MMLU} & 23.05\% & 23.58\% & 23.25\% & 23.12\% & \textbf{24.06\%} \\
\textbf{GLUE} & 45.04\% & 45.59\% & 47.94\% & 47.97\% & \textbf{48.11\%} \\
\textbf{OpenBookQA} & 15.60\% & 15.40\% & 16.80\% & 17.60\% & \textbf{17.80\%} \\
\hline
\textbf{Average} & 30.82\% & 31.04\% & 32.88\% & 33.15\% &\textbf{ 34.12\%} \\
\hline
\end{tabular}
\renewcommand{\arraystretch}{1}  % Reset to default value if needed
\label{tab:abl_b_result}

\end{table}

While the trend is monotonic, the performance gains exhibit diminishing returns relative to the increase in parameter count. As shown in Table \ref{tab:abl_b_result}, an increase in the rank from $r_c=16$ to $r_c=32$ brings about a dramatic improvement of 1.84 percentage points in average performance. In contrast, an increase in the rank to $r_c=64$ brings about a much smaller improvement of only 0.27 percentage points.

This result shows that beyond $r_c=32$, the marginal performance benefit is increasingly outweighed by the additional parameter cost, which is antithetical to the goal of compression. Therefore, the choice of $r_c=r=32$ for the main experiments is empirically supported since it achieves a trade-off between functional restoration and model efficiency. When performance is the top consideration and an increased budget for parameters is feasible, setting a higher corrective rank ($r_c > 32$) remains a viable alternative.

\subsubsection{Impact of Selective FFN Module Targeting}
\label{sssec:ablation_ffn_targeting}

A key methodological choice in CALR, outlined in Section \ref{ssec:training_optimization_calr}, is the selective application of compression to FFN modules exhibiting the lowest transformation values (as defined in (\ref{eq:transformation_metric})). To validate this, we conducted an ablation study comparing our proposed targeting strategy against two alternative approaches. All experiments were performed with a fixed number of FFN modules ($N_{\text{target}}$) compressed to ensure a fair comparison of the selection strategies themselves.

The three experimental conditions were:
\begin{enumerate}
    \item Model 1 (Lowest Transformation): This is the standard CALR approach, applying compression to the $N_{\text{target}}$ FFN modules with the lowest transformation scores.
    \item Model 2 (Highest Transformation): CALR was applied to the $N_{\text{target}}$ FFN modules with the highest transformation scores, representing the most functionally dynamic layers.
    \item Model 3 (Random Selection): CALR was applied to $N_{\text{target}}$ FFN modules chosen uniformly at random, serving as a neutral baseline.
\end{enumerate}

\begin{table}[h!]
\caption{Impact of FFN module selection strategy on performance. 
% This validates our selective compression criterion as a key design choice.
}
\centering
\renewcommand{\arraystretch}{1.3}
\begin{tabular}{p{2cm} |  c  c  c  }
\hline
\textbf{Benchmark} & \textbf{Model 1} & \textbf{Model 2} & \textbf{Model 3} \\
\hline
\textbf{ARC-C} & \textbf{26.88\%} & 24.91\% & 25.77\% \\
\textbf{ARC-E} & \textbf{44.99\%} & 27.36\% & 27.99\% \\
\textbf{HellaSwag} & \textbf{37.41\%} & 27.61\% & 28.98\% \\
\textbf{MMLU} & 23.25\% & 23.35\% & \textbf{23.37\%} \\
\textbf{GLUE} & \textbf{47.94\%} & 43.21\% & 45.02\% \\
\textbf{OpenBookQA} & \textbf{16.80\%} & 16.00\% & 14.20\% \\
\hline
\textbf{Average} & \textbf{32.88\%} & 27.07\% & 27.55\% \\
\hline
\end{tabular}
\renewcommand{\arraystretch}{1}  % Reset to default value if needed
\label{tab:abl_c_result}

\end{table}

The results in Table \ref{tab:abl_c_result} provide unequivocal support for our selective strategy. Our method (Model 1) significantly outperforms both alternatives. Targeting the most functionally dynamic layers (Model 2) proved to be the worst strategy, suggesting that their complex operations are severely disrupted by SVD. This study confirms that selectively compressing functionally stable modules is a cornerstone of CALR's effectiveness.

\subsubsection{Why Not Compress Attention (QKV) Layers with CALR?}
\label{sssec:ablation_attention_qkv}

To validate our design choice of focusing on FFNs, we conducted an ablation study extending CALR to attention QKV layers in addition to FFNs. The results in Table \ref{tab:abl_d_result} show this is significantly detrimental, causing the average performance to drop by 4.43 percentage points.

\begin{table}[h!]
\caption{Comparing CALR on FFN Only vs. FFN + QKV. 
% Extending CALR to QKV layers reduces average performance, justifying our FFN-only focus.
}
\centering
\renewcommand{\arraystretch}{1.3}
\begin{tabular}{p{2cm} |  c  c }
\hline
\textbf{Benchmark} & \textbf{CALR} & \textbf{CALR+QKV} \\
\hline
\textbf{ARC-C} & \textbf{26.88\%} & 26.71\% \\
\textbf{ARC-E} & \textbf{44.99\%} & 31.31\% \\
\textbf{HellaSwag} & \textbf{37.41\%} & 29.26\% \\
\textbf{MMLU} & \textbf{23.25\%} & 22.95\% \\
\textbf{GLUE} & \textbf{47.94\%} & 44.87\% \\
\textbf{OpenBookQA} & \textbf{16.80\%} & 15.60\% \\ 
\hline
\textbf{Average} & \textbf{32.88\%} & 28.45\% \\
\hline
\end{tabular}
\renewcommand{\arraystretch}{1}  % Reset to default value if needed
\label{tab:abl_d_result}

\end{table}

This result indicates that the CALR framework, as designed, is suboptimal for compressing QKV layers. The reasons are discussed further in Section \ref{ssec:limitations_and_scope}. This study decisively substantiates our design decision to focus CALR on FFN modules, where its two-component approach yields a highly favorable compression-performance trade-off.

\section{Discussion}
\label{sec:discussion}

\subsection{Functional Residual Correction}
\label{ssec:residual_principle}
A core question is \emph{why} modeling the functional residual with a dense, low-rank module is so effective. The answer lies in the nature of SVD-induced information loss and the objective of the correction.

\subsubsection{Architectural Congruence with SVD Information Loss} 
SVD preserves the dominant, high-variance components of a weight matrix. However, functional degradation in LLMs often stems from the cumulative loss of subtle, distributed information encoded across many smaller singular values. This lost information is not noise but a functionally critical, structured signal. Our hypothesis, supported by Fig. \ref{fig:residual_spectrum}, is that this \emph{functional residual is itself low-rank in nature}. This explains the architectural superiority of CALR's dense corrective module over a sparse alternative like that in LoSparse. A sparse matrix is designed for isolated, high-magnitude errors, which is incongruent with the distributed nature of SVD-induced functional loss. CALR's parallel, low-rank corrective module is architecturally matched to capture this distributed, low-rank residual.

\subsubsection{Optimizing for Functional over Reconstruction Fidelity} 
Standard SVD refinement methods improve the $L_2$-norm approximation of the weight matrix. CALR's training, however, shifts the objective to directly minimize the \emph{functional output error} on a calibration dataset. This allows the corrective module to learn the nuanced, non-linear ``deltas'' truncated by SVD, recovering critical behavioral capabilities not captured by a simple matrix fidelity metric.

\subsection{Architectural Limitations and Baseline Scope}
\label{ssec:limitations_and_scope}
Our study's scope and choice of baselines were carefully considered to ensure a fair and insightful evaluation of CALR's core contribution.

\subsubsection{Justification for Baseline Selection} 
To rigorously benchmark CALR, we selected a diverse set of strong baselines representing distinct compression philosophies:
\begin{itemize}
    \item LoSparse: Chosen as the \emph{most direct conceptual competitor}. Its use of a \textit{sparse} corrective term provides the perfect experimental control to test our central hypothesis that a \textit{dense, low-rank} corrective module is better suited for this context.
    \item LaCo and ShortGPT: Selected as leading examples of \emph{layer-removal techniques}. They provide a crucial contrast, testing the trade-off of reducing model \emph{depth} versus CALR's approach of reducing the internal \emph{rank} (``width") of layers.
\end{itemize}

\subsubsection{Justification for Omitted Baselines} 
We deliberately omitted other methods to maintain a focused comparison, as they address different problem facets or operate under different assumptions (e.g., matrix-level refinement like SVD-LLM \cite{wang2025svdllm}/ResSVD \cite{bai2025ressvd}, training-free compensation like EoRA \cite{eora2024shih}, or layer replacement like FlexiGPT \cite{smith2025flexigpt}), which would make direct, equitable comparison difficult.

\subsubsection{FFN-Centric Design and Task Scope}
Our most significant architectural limitation, confirmed by ablation studies (Section \ref{sssec:ablation_attention_qkv}), is that directly applying CALR to attention (QKV) layers is detrimental. This is a deliberate design choice. The non-linear softmax function in attention is highly sensitive to input distortions, and CALR's simple additive correction is structurally insufficient to repair these intricate, post-projection errors. This validates our focus on the more stable linear transformations within FFN layers.

Additionally, our current evaluation focuses on NLU benchmarks. While these tasks are comprehensive for assessing language understanding, they do not cover free-form generative capabilities. The performance of CALR on tasks like summarization, translation, or creative writing remains an open question for future work.

\section{Conclusion and Future Work}
\label{sec:conclusion_future_work}

\subsection{Conclusion}
\label{ssec:conclusion}
The deployment of LLMs is often hindered by their substantial size and computational cost. While low-rank factorization via SVD is a principled compression approach, its focus on matrix reconstruction error often leads to significant degradation in functional performance. The core challenge is that preserving mathematical fidelity does not guarantee the preservation of an LLM's nuanced capabilities.

To address this gap, we introduced CALR, a compression framework built on a two-component approach: a primary path of SVD-compressed layers, augmented by a \textit{parallel, learnable, low-rank corrective module}. This module is explicitly trained to model and compensate for the functional information lost during compression, effectively learning the ``functional residual.'' This dual-path design, which intelligently targets functionally stable FFN modules, represents an effective paradigm for preserving model fidelity under aggressive compression.

The large-scale experiments across a range of large language models (LLMs) offer empirical validation of this approach. CALR consistently delivered a better trade-off between parameter reduction and performance preservation compared to standard baselines. We demonstrated significant compression with a reduction of parameters between 27\% and 51\%, while retaining a large fraction of the original model performance between 77\% and 89\%. Our ablation experiments confirmed that the corrective module is the essential enabler for this performance recovery and that our heuristic strategy of choosing stable feedforward neural network (FFN) modules is indispensable to its success.

In conclusion, CALR's perspective of treating information loss not as an unavoidable artifact but as a learnable signal enables the creation of smaller, more efficient LLMs without the precipitous performance drops often associated with aggressive compression. This work contributes to the broader accessibility and practical deployment of powerful language models in real-world, resource-limited environments.

\subsection{Future Work}
\label{ssec:future_work}
Building on CALR's promising results, we identify several key avenues for future research.
\begin{enumerate}
    \item Adaptive and Specialized Corrective Modules: The findings of our work suggest a universal corrective module is insufficient for attention layers. Follow-up research will involve building specialized corrective modules that are tailored to the unique computational behaviors of different types of modules.
    \item Hybrid Compression Strategies: The full potential of CALR is best achieved when it is combined with other compression methods. We suggest investigating hybrid approaches that combine CALR's structured decomposition method with quantization or sparse pruning techniques.
    \item Broadening Applicability and Interpretability: A natural next step is to test the applicability of CALR to a larger range of architectures (including encoder-decoder and multimodal-based ones), languages, and operations encompassing free-text generation. At the same time, a thorough analysis regarding the corrective modules learned is necessary since it can provide valuable insights into the amount of loss encountered while compressing the information and thus inform the development of stronger and more advanced compression algorithms.
\end{enumerate}

% \section*{Acknowledgment}

%% file: a_TAI_arxiv_submission.bbl
\begin{thebibliography}{10}
\providecommand{\url}[1]{#1}
\csname url@samestyle\endcsname
\providecommand{\newblock}{\relax}
\providecommand{\bibinfo}[2]{#2}
\providecommand{\BIBentrySTDinterwordspacing}{\spaceskip=0pt\relax}
\providecommand{\BIBentryALTinterwordstretchfactor}{4}
\providecommand{\BIBentryALTinterwordspacing}{\spaceskip=\fontdimen2\font plus
\BIBentryALTinterwordstretchfactor\fontdimen3\font minus \fontdimen4\font\relax}
\providecommand{\BIBforeignlanguage}[2]{{%
\expandafter\ifx\csname l@#1\endcsname\relax
\typeout{** WARNING: IEEEtran.bst: No hyphenation pattern has been}%
\typeout{** loaded for the language `#1'. Using the pattern for}%
\typeout{** the default language instead.}%
\else
\language=\csname l@#1\endcsname
\fi
#2}}
\providecommand{\BIBdecl}{\relax}
\BIBdecl

\bibitem{brown2020language}
T.~Brown, B.~Mann, N.~Ryder, M.~Subbiah, J.~D. Kaplan, P.~Dhariwal, A.~Neelakantan, P.~Shyam, G.~Sastry, A.~Askell \emph{et~al.}, ``Language models are few-shot learners,'' \emph{Advances in neural information processing systems}, vol.~33, pp. 1877--1901, 2020.

\bibitem{openai2023gpt4}
J.~Achiam, S.~Adler, S.~Agarwal, L.~Ahmad, I.~Akkaya, F.~L. Aleman, D.~Almeida, J.~Altenschmidt, S.~Altman, S.~Anadkat \emph{et~al.}, ``Gpt-4 technical report,'' \emph{arXiv preprint arXiv:2303.08774}, 2023.

\bibitem{hagos2024}
D.~H. Hagos, R.~Battle, and D.~B. Rawat, ``Recent advances in generative ai and large language models: Current status, challenges, and perspectives,'' \emph{IEEE Transactions on Artificial Intelligence}, vol.~5, no.~12, pp. 5873--5893, 2024.

\bibitem{zhu2024surveycompression}
X.~Zhu, J.~Li, Y.~Liu, C.~Ma, and W.~Wang, ``A survey on model compression for large language models,'' \emph{Transactions of the Association for Computational Linguistics}, vol.~12, pp. 1556--1577, 2024.

\bibitem{wang2024surveyslm}
F.~Wang, Z.~Zhang, X.~Zhang, Z.~Wu, T.~Mo, Q.~Lu, W.~Wang, R.~Li, J.~Xu, X.~Tang \emph{et~al.}, ``A comprehensive survey of small language models in the era of large language models: Techniques, enhancements, applications, collaboration with llms, and trustworthiness,'' \emph{CoRR}, 2024.

\bibitem{cheng2024surveypruning}
H.~Cheng, M.~Zhang, and J.~Q. Shi, ``A survey on deep neural network pruning: Taxonomy, comparison, analysis, and recommendations,'' \emph{IEEE Transactions on Pattern Analysis and Machine Intelligence}, 2024.

\bibitem{fontana2024}
F.~Fontana, R.~Lanzino, M.~R. Marini, D.~Avola, L.~Cinque, F.~Scarcello, and G.~L. Foresti, ``Distilled gradual pruning with pruned fine-tuning,'' \emph{IEEE Transactions on Artificial Intelligence}, vol.~5, no.~8, pp. 4269--4279, 2024.

\bibitem{wang2025svdllm}
X.~Wang, Y.~Zheng, Z.~Wan, and M.~Zhang, ``Svd-llm: Truncation-aware singular value decomposition for large language model compression,'' \emph{CoRR}, 2024.

\bibitem{kaushal2021lord}
\BIBentryALTinterwordspacing
A.~Kaushal, T.~Vaidhya, and I.~Rish, ``Lo{RD}: Low-rank decomposition of monolingual code {LLM}s for one-shot compression,'' in \emph{ICML 2024 Workshop on Foundation Models in the Wild}, 2024. [Online]. Available: \url{https://openreview.net/forum?id=br49PQvuMp}
\BIBentrySTDinterwordspacing

\bibitem{Eckart_Young_1936}
C.~Eckart and G.~Young, ``The approximation of one matrix by another of lower rank,'' \emph{Psychometrika}, vol.~1, no.~3, p. 211–218, 1936.

\bibitem{MIRSKY1960}
\BIBentryALTinterwordspacing
L.~MIRSKY, ``Symmetric gauge functions and unitarily invariant norms,'' \emph{The Quarterly Journal of Mathematics}, vol.~11, no.~1, pp. 50--59, 01 1960. [Online]. Available: \url{https://doi.org/10.1093/qmath/11.1.50}
\BIBentrySTDinterwordspacing

\bibitem{GOLUB1987317}
\BIBentryALTinterwordspacing
G.~Golub, A.~Hoffman, and G.~Stewart, ``A generalization of the eckart-young-mirsky matrix approximation theorem,'' \emph{Linear Algebra and its Applications}, vol. 88-89, pp. 317--327, 1987. [Online]. Available: \url{https://www.sciencedirect.com/science/article/pii/0024379587901145}
\BIBentrySTDinterwordspacing

\bibitem{anonymous2025asvd}
Z.~Yuan, Y.~Shang, Y.~Song, Q.~Wu, Y.~Yan, and G.~Sun, ``Asvd: Activation-aware singular value decomposition for compressing large language models,'' \emph{arXiv preprint arXiv:2312.05821}, 2023.

\bibitem{sy2025lillama}
Y.~Sy, C.~Cerisara, and I.~Illina, ``Large language models compression via low-rank feature distillation,'' \emph{arXiv preprint arXiv:2412.16719}, 2024.

\bibitem{bai2025ressvd}
\BIBentryALTinterwordspacing
H.~Bai, S.~Jian, T.~Liang, Y.~Yin, and H.~Wang, ``Ressvd: Residual compensated svd for large language model compression,'' 2025. [Online]. Available: \url{https://arxiv.org/abs/2505.20112}
\BIBentrySTDinterwordspacing

\bibitem{eora2024shih}
\BIBentryALTinterwordspacing
S.-Y. Liu, H.~Yang, C.-Y. Wang, N.~C. Fung, H.~Yin, C.~Sakr, S.~Muralidharan, K.-T. Cheng, J.~Kautz, Y.-C.~F. Wang, P.~Molchanov, and M.-H. Chen, ``Eora: Training-free compensation for compressed llm with eigenspace low-rank approximation,'' \emph{CoRR}, vol. abs/2410.21271, 2024. [Online]. Available: \url{https://doi.org/10.48550/arXiv.2410.21271}
\BIBentrySTDinterwordspacing

\bibitem{li2023losparse}
Y.~Li, Y.~Yu, Q.~Zhang, C.~Liang, P.~He, W.~Chen, and T.~Zhao, ``Losparse: Structured compression of large language models based on low-rank and sparse approximation,'' in \emph{International Conference on Machine Learning}.\hskip 1em plus 0.5em minus 0.4em\relax PMLR, 2023, pp. 20\,336--20\,350.

\bibitem{hu2021lora}
E.~J. Hu, P.~Wallis, Z.~Allen-Zhu, Y.~Li, S.~Wang, L.~Wang, W.~Chen \emph{et~al.}, ``Lora: Low-rank adaptation of large language models,'' in \emph{International Conference on Learning Representations}, 2022.

\bibitem{men2024shortgpt}
X.~Men, M.~Xu, Q.~Zhang, B.~Wang, H.~Lin, Y.~Lu, X.~Han, and W.~Chen, ``Shortgpt: Layers in large language models are more redundant than you expect,'' \emph{CoRR}, 2024.

\bibitem{yang2024laco}
Y.~Yang, Z.~Cao, and H.~Zhao, ``Laco: Large language model pruning via layer collapse,'' in \emph{Findings of the Association for Computational Linguistics: EMNLP 2024}, 2024, pp. 6401--6417.

\bibitem{michel2019sixteenheads}
P.~Michel, O.~Levy, and G.~Neubig, ``Are sixteen heads really better than one?'' \emph{Advances in neural information processing systems}, vol.~32, 2019.

\bibitem{ma2023llmpruner}
X.~Ma, G.~Fang, and X.~Wang, ``Llm-pruner: On the structural pruning of large language models,'' \emph{Advances in neural information processing systems}, vol.~36, pp. 21\,702--21\,720, 2023.

\bibitem{ashkboos2024slicegpt}
S.~Ashkboos, M.~L. Croci, M.~G. do~Nascimento, T.~Hoefler, and J.~Hensman, ``Slicegpt: Compress large language models by deleting rows and columns,'' in \emph{The Twelfth International Conference on Learning Representations}, 2024.

\bibitem{allal2025smollm2}
L.~B. Allal, A.~Lozhkov, E.~Bakouch, G.~M. Bl{\'a}zquez, G.~Penedo, L.~Tunstall, A.~Marafioti, H.~Kydl{\'\i}{\v{c}}ek, A.~P. Lajar{\'\i}n, V.~Srivastav \emph{et~al.}, ``Smollm2: When smol goes big--data-centric training of a small language model,'' \emph{arXiv preprint arXiv:2502.02737}, 2025.

\bibitem{yang2025qwen3}
A.~Yang, A.~Li, B.~Yang, B.~Zhang, B.~Hui, B.~Zheng, B.~Yu, C.~Gao, C.~Huang, C.~Lv \emph{et~al.}, ``Qwen3 technical report,'' \emph{arXiv preprint arXiv:2505.09388}, 2025.

\bibitem{meta2024llama1b}
{Meta}, ``{Llama-3.2-1B},'' \url{https://huggingface.co/meta-llama/Llama-3.2-1B}, 2024, accessed: 2025-06-06.

\bibitem{meta2024llama3b}
------, ``{Llama-3.2-3B},'' \url{https://huggingface.co/meta-llama/Llama-3.2-3B}, 2024, accessed: 2025-06-06.

\bibitem{zhang2024layerimportance}
Y.~Zhang, Y.~Dong, and K.~Kawaguchi, ``Investigating layer importance in large language models,'' in \emph{Proceedings of the 7th BlackboxNLP Workshop: Analyzing and Interpreting Neural Networks for NLP}, 2024, pp. 469--479.

\bibitem{rajapaksha2024}
P.~Rajapaksha and N.~Crespi, ``Explainable attention pruning: A metalearning-based approach,'' \emph{IEEE Transactions on Artificial Intelligence}, vol.~5, no.~6, pp. 2505--2516, 2024.

\bibitem{le2024cosmopedia}
\BIBentryALTinterwordspacing
L.~Ben~Allal, A.~Lozhkov, G.~Penedo, T.~Wolf, and L.~von Werra, ``Cosmopedia,'' 2024. [Online]. Available: \url{https://huggingface.co/datasets/HuggingFaceTB/cosmopedia}
\BIBentrySTDinterwordspacing

\bibitem{hendrycks2021measuring}
D.~Hendrycks, C.~Burns, S.~Basart, A.~Zou, M.~Mazeika, D.~Song, and J.~Steinhardt, ``Measuring massive multitask language understanding,'' in \emph{International Conference on Learning Representations}, 2021.

\bibitem{zellers2019hellaswag}
R.~Zellers, A.~Holtzman, Y.~Bisk, A.~Farhadi, and Y.~Choi, ``Hellaswag: Can a machine really finish your sentence?'' in \emph{Proceedings of the 57th Annual Meeting of the Association for Computational Linguistics}, 2019, pp. 4791--4800.

\bibitem{clark2018think}
P.~Clark, I.~Cowhey, O.~Etzioni, T.~Khot, A.~Sabharwal, C.~Schoenick, and O.~Tafjord, ``Think you have solved question answering? try arc, the ai2 reasoning challenge,'' \emph{arXiv preprint arXiv:1803.05457}, 2018.

\bibitem{mihaylov2018can}
T.~Mihaylov, P.~Clark, T.~Khot, and A.~Sabharwal, ``Can a suit of armor conduct electricity? a new dataset for open book question answering,'' in \emph{Proceedings of the 2018 Conference on Empirical Methods in Natural Language Processing}, 2018, pp. 2381--2391.

\bibitem{wang2018glue}
A.~Wang, A.~Singh, J.~Michael, F.~Hill, O.~Levy, and S.~R. Bowman, ``Glue: A multi-task benchmark and analysis platform for natural language understanding,'' in \emph{International Conference on Learning Representations}, 2019.

\bibitem{loshchilov2017decoupled}
\BIBentryALTinterwordspacing
I.~Loshchilov and F.~Hutter, ``Decoupled weight decay regularization,'' in \emph{7th International Conference on Learning Representations, {ICLR} 2019, New Orleans, LA, USA, May 6-9, 2019}.\hskip 1em plus 0.5em minus 0.4em\relax OpenReview.net, 2019. [Online]. Available: \url{https://openreview.net/forum?id=Bkg6RiCqY7}
\BIBentrySTDinterwordspacing

\bibitem{gao2021framework}
\BIBentryALTinterwordspacing
L.~Gao, J.~Tow, B.~Abbasi, S.~Biderman, S.~Black, A.~DiPofi, C.~Foster, L.~Golding, J.~Hsu, A.~Le~Noac'h, H.~Li, K.~McDonell, N.~Muennighoff, C.~Ociepa, J.~Phang, L.~Reynolds, H.~Schoelkopf, A.~Skowron, L.~Sutawika, E.~Tang, A.~Thite, B.~Wang, K.~Wang, and A.~Zou, ``The language model evaluation harness,'' 07 2024. [Online]. Available: \url{https://zenodo.org/records/12608602}
\BIBentrySTDinterwordspacing

\bibitem{wolf2020transformers}
\BIBentryALTinterwordspacing
T.~Wolf, L.~Debut, V.~Sanh, J.~Chaumond, C.~Delangue, A.~Moi, P.~Cistac, T.~Rault, R.~Louf, M.~Funtowicz, J.~Davison, S.~Shleifer, P.~von Platen, C.~Ma, Y.~Jernite, J.~Plu, C.~Xu, T.~Le~Scao, S.~Gugger, M.~Drame, Q.~Lhoest, and A.~Rush, ``Transformers: State-of-the-art natural language processing,'' in \emph{Proceedings of the 2020 Conference on Empirical Methods in Natural Language Processing: System Demonstrations}, Q.~Liu and D.~Schlangen, Eds.\hskip 1em plus 0.5em minus 0.4em\relax Online: Association for Computational Linguistics, Oct. 2020, pp. 38--45. [Online]. Available: \url{https://aclanthology.org/2020.emnlp-demos.6/}
\BIBentrySTDinterwordspacing

\bibitem{Friedman01121937}
\BIBentryALTinterwordspacing
M.~F. and, ``The use of ranks to avoid the assumption of normality implicit in the analysis of variance,'' \emph{Journal of the American Statistical Association}, vol.~32, no. 200, pp. 675--701, 1937. [Online]. Available: \url{https://www.tandfonline.com/doi/abs/10.1080/01621459.1937.10503522}
\BIBentrySTDinterwordspacing

\bibitem{wilcoxon1945}
\BIBentryALTinterwordspacing
F.~Wilcoxon, ``Individual comparisons by ranking methods,'' \emph{Biometrics Bulletin}, vol.~1, no.~6, pp. 80--83, 1945. [Online]. Available: \url{http://www.jstor.org/stable/3001968}
\BIBentrySTDinterwordspacing

\bibitem{holm1979simple}
S.~Holm, ``A simple sequentially rejective multiple test procedure,'' \emph{Scandinavian Journal of Statistics}, vol.~6, no.~2, pp. 65--70, 1979.

\bibitem{smith2025flexigpt}
J.~S. Smith, C.-H. Lin, S.~Tuli, H.~Jeelani, S.~Gao, Y.~Shen, H.~Jin, and Y.-C. Hsu, ``Flexigpt: Pruning and extending large language models with low-rank weight sharing,'' \emph{arXiv preprint arXiv:2501.14713}, 2025.

\end{thebibliography}
